\documentclass[10pt,twocolumn,letterpaper]{article}
\usepackage{cvpr}              

\usepackage{color, colortbl}
\usepackage{multirow}

\definecolor{tabfirst}{rgb}{1, 0.7, 0.7} 
\definecolor{tabsecond}{rgb}{1, 0.85, 0.7} 
\definecolor{tabthird}{rgb}{1, 1, 0.7} 

\definecolor{cvprblue}{rgb}{0.21,0.49,0.74}
\usepackage[pagebackref,breaklinks,colorlinks,allcolors=cvprblue]{hyperref}

\def\paperID{14224} 
\def\confName{CVPR}
\def\confYear{2025}

\title{ LookCloser: Frequency-aware Radiance Field for Tiny-Detail Scene}

\author{
  Xiaoyu Zhang$^{1}$\footnotemark[1]\and
  Weihong Pan$^{2}$\footnotemark[1]\and 
  Chong Bao$^{2}$\and
  Xiyu Zhang$^{2}$ \and 
  Xiaojun Xiang$^{1}$\and
  Hanqing Jiang$^{1}$\footnotemark[2]\and
  Hujun Bao$^{2}$\footnotemark[2]\and
  \textnormal{\hspace{-1em}$^1$SenseTime Research \quad $^2$State Key Lab of CAD\&CG, Zhejiang University}
} 

\begin{document}
\twocolumn[{%
\renewcommand\twocolumn[1][]{#1}%
\maketitle
\centering
\vspace{-0.5cm}
\includegraphics[width=0.99\linewidth]{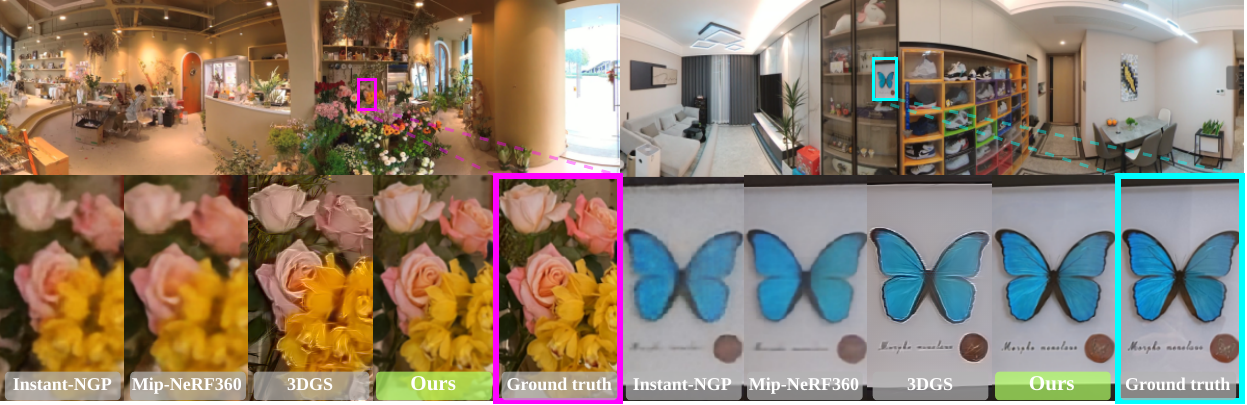}
\captionof{figure}{\textbf{FA-NeRF} is a frequency-aware framework to capture both the overall scene structure and the high-definition tiny details within a single model, \eg, the clear petal of flowers and sharp texture of wings of the butterfly. Instead, other methods give an over-smooth rendering and lose excessive details. Please see the \textbf{Supplementary Video} for an immersive roaming experience.}
\label{fig:figure1}
\vspace{4mm}
}]
\renewcommand{\thefootnote}{\fnsymbol{footnote}}
\footnotetext[1]{Authors contributed equally.}
\footnotetext[2]{Corresponding authors.}
\begin{abstract}
Humans perceive and comprehend their surroundings through information spanning multiple frequencies. In immersive scenes, people naturally scan their environment to grasp its overall structure while examining fine details of objects that capture their attention. However, current NeRF frameworks primarily focus on modeling either high-frequency local views or the broad structure of scenes with low-frequency information, which is limited to balancing both.
We introduce FA-NeRF, a novel frequency-aware framework for view synthesis that simultaneously captures the overall scene structure and high-definition details within a single NeRF model. To achieve this, we propose a 3D frequency quantification method that analyzes the scene’s frequency distribution, enabling frequency-aware rendering. Our framework incorporates a frequency grid for fast convergence and querying, a frequency-aware feature re-weighting strategy to balance features across different frequency contents. Extensive experiments show that our method significantly outperforms existing approaches in modeling entire scenes while preserving fine details.
\end{abstract}
    
\section{Introduction}
\label{sec:intro}
The Neural Radiance Field (NeRF) has achieved significant success in photo-realistic novel view synthesis and shows great potential for immersive roaming experiences, game and film production, etc. Significant efforts have been invested to enhance the performance of NeRF~\cite{barron2021mip,Reiser_2021_ICCV,huang2023local}, including improving the high-frequency details~\cite{Zhou_2023_CVPR,Xu_2023_CVPR,Li_2023_ICCV,turki2024pynerf}of localized scenes or adapting to the expanded scale of scenes~\cite{turki2022mega,tancik2022block,Isaac-Medina_2023_CVPR}.

However, in real-world scenarios, both scene structure and intricate details are crucial. Artifacts from incomplete structures or excessive blurring of details can significantly undermine the sense of immersion. To address this, we constructed a dataset—termed the Multi-frequency Dataset—that captures panoramic images for low-frequency structural information and high-resolution images for high-frequency details.

In this work, we aim to provide a high-quality rendering of both scene structures and intricate details across a wide frequency range. However, due to varying viewpoints and resolutions in the dataset, frequency variations in the captured 3D signals can differ by the order of magnitude. These extensive frequency variations present substantial challenges for NeRF and its derivatives. While methods such as Mip-NeRF 360~\cite{barron2022mip} incorporate scale representations by casting cones instead of rays to achieve anti-aliasing and high-quality rendering, they perform unsatisfactorily when reconstructing multi-frequency signals together~\cite{dhiman2023strata}. This limitation arises from the uniform treatment of pixels, which overlooks the frequency distribution within the scene.
Other approaches, such as BungeeNeRF~\cite{xiangli2022bungeenerf} and Strata~\cite{dhiman2023strata}, attempt to address large viewpoint variations by progressively enabling high-frequency feature components to capture fine details, but they struggle to generalize in more varied settings. Adaptive spatial partitioning~\cite{saragadam2022miner, wang2023f2} is another common strategy, designed to selectively capture scene content at different frequencies. However, partitioning based solely on spatial relationships may not align with the actual frequency distribution, potentially limiting the representation of high-frequency details.

To address these challenges, we propose FA-NeRF, a frequency-aware framework for high-fidelity novel view synthesis across broad scene structures and close-up details. FA-NeRF introduces a patch-based 3D frequency quantification method to analyze and embed scene frequency distributions into NeRF’s encoded features, allowing adaptive frequency selection for accurate reconstruction.

Efficiently synthesizing high-quality views with detailed structure remains challenging even with frequency quantification. First, we design a Frequency Grid that stores spatial frequency distributions, enabling rapid convergence and efficient querying. Second, we propose a frequency-aware feature re-weighting strategy to tailor feature frequencies based on scene content, optimizing network capacity. Third, a frequency-averaged sampling strategy adaptively adjusts learning intensity and sampling density for high-frequency content. With the Hash Grid architecture, we maintain a rendering speed of 20 FPS on a single RTX 4090 GPU, even when rendering high-frequency details. In summary, our contributions lie in three aspects:
\begin{itemize}
   \item The proposed FA-NeRF framework features both the scene's overall structure and tiny details within a single model, achieving an immersive roaming experience in rendering with large frequency spans.
   \item A novel patch-based 3D frequency quantification method using image progressive regression and conducting several novel techniques: a frequency grid for fast frequency convergence and query, feature re-weighting, and sampling adjustment to enhance the model's sensitivity to various frequency content. 
   \item FA-NeRF significantly outperforms the baselines across our Multi-frequency dataset and generalizes well on 2 standard datasets.
\end{itemize}

\begin{figure*}[t]       
       \centering
       \includegraphics[width=0.98\linewidth]{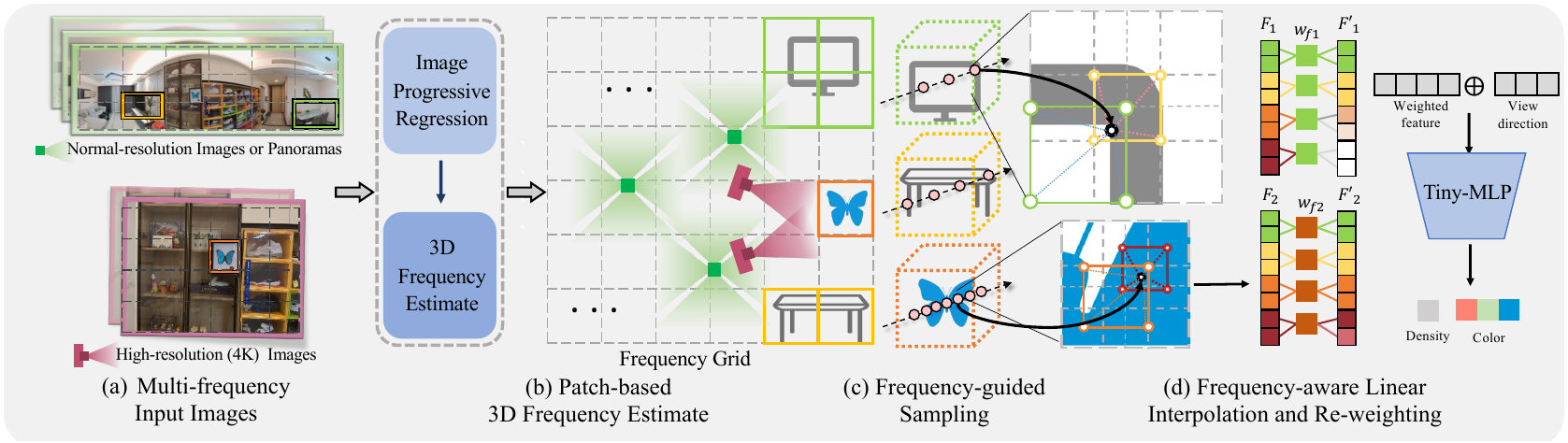}
    
           \caption{\textbf{Overview}. (a) Our input dataset consists of dense normal-resolution images or panoramas of the entire scene structure and high-resolution images focusing on the area of interest of the scene. (b) We propose to quantify the 3D frequency of the scene by progressive image regression and maintain a frequency grid to divide the scene into subspaces with different frequency distributions. (c) We employ a frequency-guided sampling to adaptively control the density of sampling according to the frequency of different objects. (d) Then, the grid-encoded feature of each sample point is re-weighted by the frequency-related weight function. The density and color are decoded by a tiny multilayer perceptron which takes the re-weighed feature and view direction as input.}
           \label{fig:pipline_v1.pdf}
           \vspace{-0.5em}
\end{figure*}    
\section{Related Work}
\label{sec:relatedwork}
\noindent\textbf{Neural scene representations.} The classical work of Mildenhall Ben et al.~\cite{mildenhall2020nerf} laid the foundation for the development of NeRF technology, but turns out to be more suitable for reconstructing an object or a bounded small-scale scene due to its huge sampling and Multi-Layer Perception (MLP) access time of training and rendering~\cite{MildenhallSTBRN20, sitzmann2020implicit, fathony2020multiplicative, lindell2022bacon, ramasinghe2022beyond}.
Early NeRF works attempted to accelerate the MLP query process by utilizing sparser geometry representation like sparse voxel grid~\cite{liu2020neural,hedman2021baking,YuFTCR22,ChenXGYS22,SunSC22}, octrees~\cite{yu2021plenoctrees,fridovich2022plenoxels}, 4D tensor~\cite{chen2022tensorf,FlynnBDDFOST19,LiXDS20,MildenhallSCKRN19} or grid mesh~\cite{chen2023mobilenerf,WaechterMG14,WoodAACDSS00,DebevecTM96,ThiesZN19}, combined with more efficient training and rendering schemes~\cite{takikawa2021neural,martel2021acorn,mueller2022instant,LiuGLCT20}.
Instead of using a huge MLP network, Instant-NGP~\cite{muller2022instant} and TensoRF~\cite{chen2022tensorf} establish a voxel grid to store features followed by a tiny MLP network for efficient training and inference. 
Recently, 3D Gaussian Splatting (3DGS)~\cite{kerbl20233dgs} and its variants ~\cite{cheng2024gaussianpro3dgaussiansplatting,xiong2024sparsegsrealtime360degsparse,zhu2024gsror3dgaussiansplatting,bulò2024revisingdensificationgaussiansplatting,li2024geogaussiangeometryawaregaussiansplatting,Yu2024MipSplatting}  has achieved great success in fast and efficient novel view synthesis by replacing NeRF’s implicit networks with explicit 3D Gaussians and accelerating rendering through differentiable rasterization. However, most of these works are designed to learn a default frequency space without fully considering the frequency distribution within the scene, which limits their ability to effectively capture structures and details in scenarios with significant frequency variations.

\noindent\textbf{Multi-scale Scene Representation.} 
Multi-scale representation~\cite{barron2021mip, lindell2022bacon, xiangli2022bungeenerf, dhiman2023strata, hu2023tri,yariv2020multiview, wang2021neus} propose novel grid-based or anti-aliased representations to preserve multi-scale appearance and details in the trained NeRF model. 
A typical work is Mip-NeRF~\cite{barron2021mip} which proposes an anti-aliased multi-scale representation by incorporating cone casting and integrated positional encoding (IPE) features into a single multiscale MLP. Mip-NeRF-360~\cite{barron2022mip} extends Mip-NeRF from object to unbounded scene. Following Mip-NeRF-360~\cite{barron2022mip} and Instant-NGP~\cite{muller2022instant} reduces the aliasing issue and accelerates the training time by combining a multisampling-like solution with grid-based NeRF model. Some works~\cite{xiangli2022bungeenerf, dhiman2023strata}  also focus on training multi-scale scenes by progressively opening high-frequency feature components to learn the details.
Although these methods have made effective multi-scale improvements, most of them haven't learned the explicit 3D distribution of frequencies, which somehow restricts their 
representation capability for 
frequency-variant training data. 
  
\section{Preliminaries}
Neural Radiance Field (NeRF) parameterizes the scene as a continuous implicit function $F$, mapping a 3D position $\mathbf{x}\in \mathbb{R}^3 $ and a viewing direction $\mathbf{d}\in \mathbb{S}^2 $ to a color vector $\mathbf{c}\in [0,1]^3$ and a volumetric density $\sigma\in\mathbb{R}^+$. Each pixel on the image determines a ray emitted from the camera center of projection to the pixel. Instead of sending position $\mathbf{x}$ and direction $\mathbf{d}$ to the network, a positional encoding function is used to map them into a higher dimensional space. 
This can be formulated as
\begin{align}
    (\mathbf{c},\sigma ) & = F_{\theta}(\gamma_x(\mathbf{x}),\gamma_d(\mathbf{d})),
    \label{eq:nerf}
\end{align}
where $F$ denotes an MLP with parameters $\theta$, and $\gamma:\mathbb{R} ^3\to \mathbb{R} ^{3(1+2L)}$ a positional encoding function with $L$ frequency channels.
The network is then optimized following the volume rendering procedure to represent scenes with photo-realistic rendering.

\noindent\textbf{Frequency components in positional encoding.} There are mainly two types of positional encoding. Vanilla NeRF uses Fourier-transformed features to encode the position, and higher Fourier series terms correspond to higher frequency components. 
NeRF~\cite{mildenhall2020nerf} uses a simple concatenation of sines and cosines as a positional encoding function, which is applied to each dimension of the normalized 3D position $\mathbf{x}$ separately:

\begin{small}
\begin{align}
    \gamma(\mathbf{x})=(&\sin \left(2^{0} \pi \mathbf{x}\right), \cos \left(2^{0} \pi \mathbf{x}\right), \dots, \notag \\
    &\sin \left(2^{L-1} \pi \mathbf{x}\right), \cos \left(2^{L-1} \pi \mathbf{x}\right)).
\end{align}
\end{small}
$L$ determines the highest sampling rate, hence having a critical impact on the fidelity of NeRF. 

To remedy the aliasing issue caused by multiscale training data, Mip-NeRF~\cite{barron2021mip} considers a ray as a cone and divides it into several conical frustums whose mean and variance $(\mu,\Sigma)$ are used for integrated position encoding (IPE):
\begin{align}
    \gamma(\mu,\Sigma) & = \left \{ \begin{bmatrix}
 \operatorname{sin}(2^{l}\mu)\exp(-2^{2l-1}\operatorname{diag}(\Sigma))\\
 \operatorname{cos}(2^{l}\mu)\exp(-2^{2l-1}\operatorname{diag}(\Sigma))
\end{bmatrix}  \right \} _0^{L-1}.
\label{eq:positional}
\end{align}

In grid-based NeRF, parametric encoding is common. Instant-NGP (iNGP)~\cite{muller2022instant} introduces multi-resolution hash encoding, replacing positional encoding with a pyramid grid that spans coarse to fine resolutions. Each of the $L$ resolution levels in the hash table stores $F$-dimensional feature vectors at grid corners, and each 3D position $\mathbf{x}$ retrieves a feature vector by interpolating and concatenating features from surrounding corners across levels. The $L$ levels in hash encoding are analogous to frequency channels in frequency encoding: higher resolution levels capture high-frequency components, while lower resolution levels capture low-frequency components.

\begin{figure*}[t]
       \centering
       \includegraphics[width=0.98\linewidth]{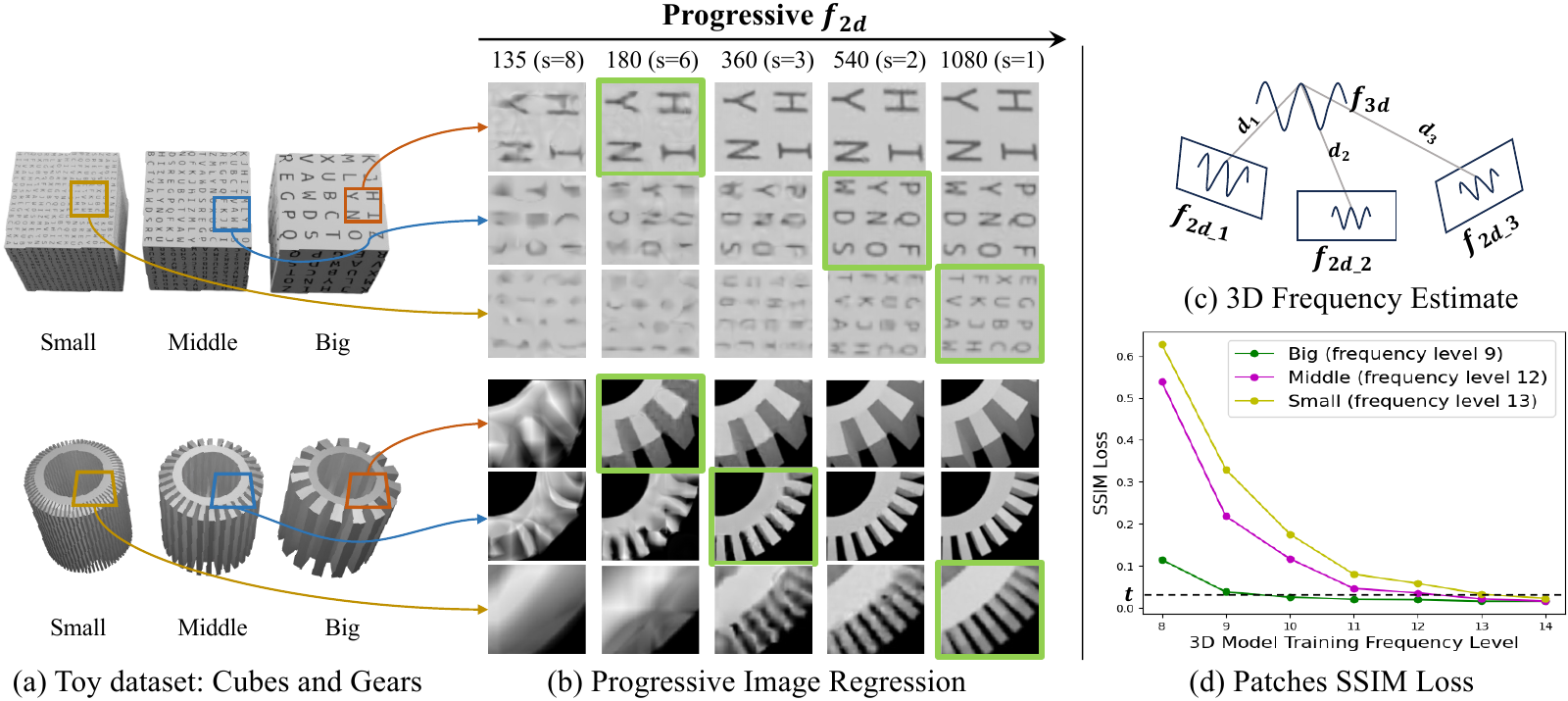}

       \caption{We show 3D frequency quantification on toy examples. (a) The toy datasets comprise three kinds of frequency variations in geometry or texture: three cubes with the appearance of different font sizes, and three gears with the geometry of different pitches. (b) The rendered high-frequency details under different frequencies. "135($s$=8)" means the frequency is 135 and the grid stride $s$ is 8~\cite{muller2022instant}.  The green squares denote the minimal frequency (selected) of the patch for clear reconstruction. (c) The 3D frequency comes from a set of projected 2D frequencies with focal length and depth. (d) The SSIM loss (vertical axis) of training results when using different frequencies (horizontal axis) in three kinds of patches. }
       \label{fig:toydatasets.pdf}
       \vspace{-1.0em}
\end{figure*} 

\section{Method}
\label{sec:blind}

We introduce FA-NeRF, a frequency-aware neural radiance field for high-fidelity novel view synthesis with multi-frequency details (see Fig.~\ref{fig:pipline_v1.pdf}). Our input images include panoramic shots from a standard camera and high-resolution (up to 4K) images from an SLR camera, with camera poses recovered via structure from motion (SfM). In Sec.~\ref{ssec:method_quantify}, we estimate the scene’s 3D frequency distribution using a patch-based quantification method. Then, in ~\ref{ssec:method_frequencyaware}, we apply a frequency-aware training framework to preserve these details, utilizing a frequency grid and re-weighting features based on the estimated frequency information.

\subsection{Evaluate Frequency Level from 2D to 3D}
\label{ssec:method_quantify}

Inspired by multi-plane image features for novel view synthesis~\cite{wizadwongsa2021nex}, we propose a hypothesis: In the NeRF framework, the geometric or appearance frequency of 3D content can be inferred from the frequency in its degraded 2D image space. 
To provide a more intuitive description of this process,  
we created a toy dataset shown in \cref{fig:toydatasets.pdf} (a), containing three cubes with the appearance of different-sized letters and three gears with the geometry of different-sized tooth pitches. These two scenes correspond to three different texture frequencies and geometric frequencies respectively.

\noindent\textbf{Progressive image regression.}  
Our key idea is to progressively add higher-frequency encoded feature components of NeRF until the image patch recovers clear structural information.
We define this frequency as the 2D frequency of this patch.
For a NeRF network in Eq.~(\ref{eq:nerf}), we use a coordinate-based MLP network to perform image regression: $F_{\theta} (\mathbf{x})=\mathbf{c}$, where $\mathbf{x}$ is the 2D coordinate of a sampled pixel and $\mathbf{c}$ is the color.
Given a 3D point $p$ and its corresponding patch $P$, set $\left \{\hat{P}_{f_1}, \hat{P}_{f_2}, ..., \hat{P}_{f_n}\right\}$ represents the rendering results at 2D frequency $f_i$ where $i$ indexes the frequency components and ranges from  $1$ to $n$. The target 2D frequency $f_{2D}$ is defined as the minimum frequency $f$ that satisfies $SSIM(P, \hat{P}_{f})>t $. SSIM denotes Structural Similarity Index Measure to determine whether the patch fitting meets the required standards, $f$ lies in $\left\{f_1, f_2,..., f_n\right\}$, and $t$ is a predefined threshold.
As depicted in \cref{fig:toydatasets.pdf}(b), with frequency $f$ increasing from 135 to 1080, the rendering quality gradually improves. The green box indicates the first patch that satisfies the SSIM loss threshold, and the corresponding frequency is the 2D frequency of the 2D image.

\noindent\textbf{3D frequency estimation.}
For point $p$ and patch $P$ mentioned above, we project the 2D frequency $f_{2D}$ to 3D space with the focal length $fl$ and the depth $d$ of the point to get its 3D frequency
\begin{align}
f_{3D}(fl, d) = f_{2D}\cdot \frac{fl}{d}.
\label{eq:fre3d}
\end{align}
Since point $p$ has multiple visual patches, its 3D frequency set can be defined as $F=\left\{f_{3D_{j}} | j=1...,m \right\}$ calculated with patches $\left\{P_j\right\}$ and the 3D frequency of point $p$ is defined as the median of set $F$, as illustrated in Fig.~\ref{fig:toydatasets.pdf} (c).

We only perform projection in regions having 2D-3D correspondence because consistency observation is the key to generating coherent 3D content. During the training process, we update the depth with:
\begin{align}
d(\mathbf{r} ) & = \int_{t_n}^{t_f} T(t)\cdot\sigma(\mathbf{r}(t) )\cdot t\cdot dt,
\end{align}
where $t_n$ and $t_f$ denote the nearest and the farthest distance from the camera center along the ray respectively, and $T(t)$ denotes the accumulated transmittance from $t_n$ to $t$. In \cref{fig:toydatasets.pdf} (c), we show the frequency projection process from 2D to 3D space. 

Moreover, as depicted in Fig.~\ref{fig:toydatasets.pdf}(d), green, magenta, and yellow lines represent patches with 3D frequency levels of 9, 12, and 13, respectively. As the training frequency level increases from 8 to 14, the SSIM loss of rendered patches gradually decreases. When the training frequency reaches the corresponding 3D frequency of each patch, the SSIM loss drops below the threshold. This indicates that: 1) The minimum NeRF frequency level required to fully restore the structures and the textures of different 3D frequencies in the scene varies; 2) Our 3D frequency estimation for the 3D contents accurately reflects their true frequencies.

This conclusion is helpful for understanding rendering performance in multi-frequency scenes.

\begin{figure*}[t]
       \centering
       \includegraphics[width=0.98\linewidth]{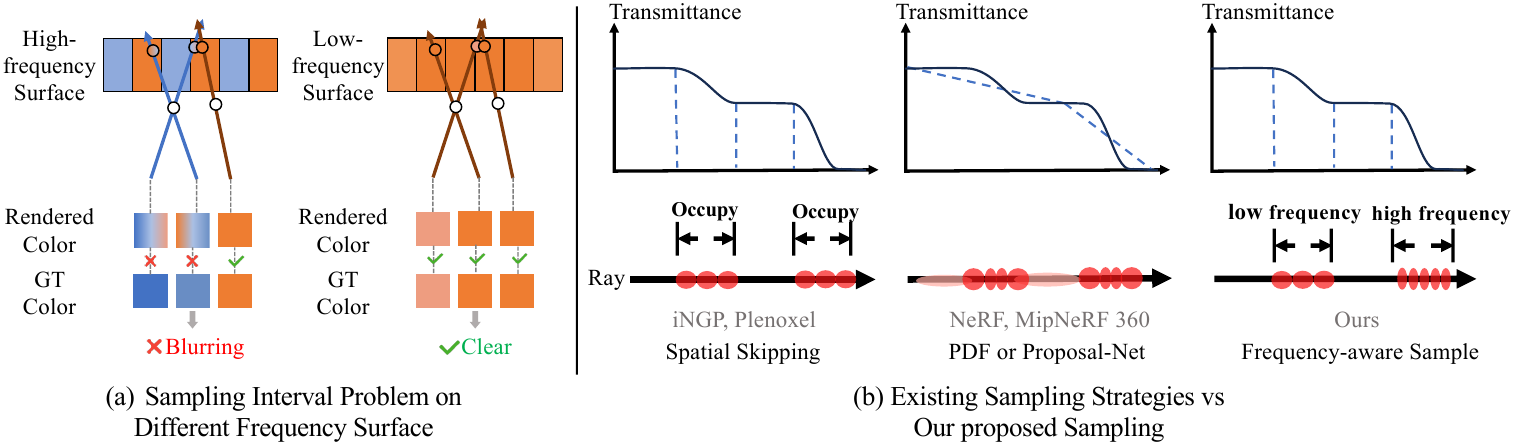}

       \caption{(a) We illustrate the sampling interval problem on different frequency surfaces. While using larger sampling intervals still achieves correct results on the low-frequency surface, it leads to misleading results on the high-frequency surface, resulting in excessive smoothing (b) We compare three sampling strategies along the ray. Our strategy adjusts the sampling interval to match the content's frequency.
       }       
       \label{fig:sample.pdf}
\end{figure*}

\subsection{Frequency-aware Framework}
\label{ssec:method_frequencyaware}

Directly training on multi-level frequency data may lead to incorrect geometry and make convergence difficult ~\cite{xiangli2022bungeenerf, dhiman2023strata}. "Coarse-to-fine" is a widely adopted learning strategy ~\cite{sun2022direct,fridovich2022plenoxels,wang2022nerf}, which first reconstructs the scene structure using low-frequency feature components and then promotes detail recovery using high-frequency feature components. However, in complex multi-frequency data, these strategies cannot accurately restore details without knowing the frequency hierarchy of the target objects. 
Moreover, as shown in Fig.~\ref{fig:sample.pdf} (a), the sampling strategy affects the rendering results in different frequency surfaces. 
Therefore, we propose a frequency grid to store the frequency distribution of the scene and adjust the NeRF-encoded features at different frequency levels through re-weighting to make more efficient use of NeRF-encoded feature space for various frequency content and adjust the sampling strategy to enhance learning high-frequency content.

\noindent\textbf{Frequency grid.} We use a frequency voxel grid $\boldsymbol{V}^{(\text{frequency})}\in\mathbb{R} ^{N_x \times N_y \times N_z \times 1}$ to store spatial occupancy information and records the frequency information of the content occupying the space, as illustrated in \cref{fig:pipline_v1.pdf}(b). $\boldsymbol{V}^{(\text{frequency})}$ is initialized by the point cloud. 
Given a 3D point $p$ in the point cloud and its observation images, we re-project it to these images and set $\left\{ P_{i} | i=1,...,n\right\}$ denotes $n$ corresponding patches where the re-projected point locates. $n$ equals the number of observation images of point $p$. 
As mentioned in Section~\ref{ssec:method_quantify}, each point $p$ has its 3D frequency $f_{{3D}}$.
Thus the $\boldsymbol{V}^{(\text{frequency})}$ is initialized with $f_{grid} = \max\left\{f_{{3D}_j}|j=1,...,m\right\}$ where $m$ denotes the amount of 3D points in this grid.
The values stored in the $\boldsymbol{V}^{(\text{frequency})}$ are normalized according to the scale of the scene to ensure consistency with the encoded feature frequency components of NeRF.

As we gradually reconstruct the scene structure and get the depth of the training ray, the frequency value is updated using Eq.~(\ref{eq:fre3d}). Since all the 2D frequencies ${f_{2D}}$ have been obtained, this process consumes only a negligible amount of computational resources. 

\noindent\textbf{Frequency re-weighting.} To achieve a better balance of features across varying frequencies, we re-weight the feature at each frequency level based on the quantified frequency in $\boldsymbol{V}^{(\text{frequency})}$. Although high-frequency feature components contribute little to low-frequency content, directly decomposing features to separately learn multi-frequency content is not effective~\cite{xiangli2022bungeenerf}. Therefore, by applying smooth re-weighting, we adjust the sensitivity of each frequency component in the NeRF-encoded features, thereby preventing wasting the expressive capacity of high-frequency feature components on low-frequency content.
In Instant-NGP~\cite{muller2022instant}, the sampled point, whose spatial location is $\mathbf{x}$, will be first scaled by the grid linear size $n_\ell$ at the frequency (\ie level) $\ell$. 
The feature of the sampled point at the frequency $\ell$ comes from tri-linear interpolation within hash grid $V_\ell$.
These vectors are directly concatenated to form the encoded feature $\mathbf{f}$. 
However, we do not directly concatenate these vectors; instead, we apply a frequency-related down-weighting factor:
\begin{align}
\omega_{\ell}=\text{erf}(\sqrt{\frac{(\ell_{\text{max}}-\ell_{\text{min}})^2}{\text{Clip}[(\ell_{\text{max}}-\ell+1)^2, \{1, (\ell_{\text{max}}-\ell_{\text{min}})^2\}]}}),
\end{align}
where $\ell_{\text{min}}$ and $\ell_{\text{max}}$  are the minimum and maximum frequency in $\boldsymbol{V}^{(\text{frequency})}$ respectively, and $\omega_{\ell}\in[0,1]$, before that:
\begin{align}
\mathbf{f}=\underset{\ell=0..k}{\text{concat}}(\omega_{\ell}\cdot\text{trilerp}(n_\ell\cdot\mathbf{x};V_\ell ) ),
    \end{align}

where $k$ is the number of levels in the hash grid $V$.
Since low-frequency feature components are used in learning high-frequency content, we opt for a one-sided weight reduction function.

\noindent\textbf{Frequency-averaged sampling.} 
Since the complexity of geometry or texture in the high-frequency area is more than that of the low-frequency area, the high-frequency area deserves more extensive learning.
However, random pixel sampling within the dataset uses a uniform probability for low-frequency and high-frequency areas in a training batch.
Therefore, we propose a frequency-averaged sampling (FAS) strategy. 
Assume there are $N$ effective frequencies in the scene and each frequency contains a patch set of size $n$($n > 0$). 
We evenly divide a training batch into $N$ segments
. Each segment samples pixels within a corresponding frequency patch,
thereby increasing the sampling probability of high-frequency areas.
To achieve a high-quality novel view synthesis, we don't sample the entire patch; instead, we adjust the probability of pixels being sampled on a patch-by-patch basis. 
We progressively employ this sampling strategy to encourage the network to form the correct geometry first. 
For more details, please refer to the supplementary material.

\noindent\textbf{Adaptive Ray Marching.} The sampling interval in ray marching affects the quality of high-frequency details, especially when reconstructing high-frequency details at the scale of the entire scene. Recall that NeRF models a 3D scene using a rendering function as Eq.~\ref{eq:nerf}, which maps the coordinates of a 3D point to the properties of the scene. 

The ray sampling interval refers to the distance between adjacent sample points along a ray. As shown in \cref{fig:sample.pdf}(a), with a large sampling interval, the low-frequency surfaces can recover their original colors, while high-frequency areas are prone to rendering incorrect colors due to the sampling points too far from the surface, resulting in blurred outcomes. When the sampling interval is reduced, the sampling points in the high-frequency areas are closer to the surface, thus generating more accurate results. 

Therefore, many studies adjust the sampling interval by manually tuning the sampling steps or setting different proposal steps for different scenes to achieve optimal performance.
We adjust the sampling interval based on the quantified frequency grid to accommodate the sampling needs of high-frequency areas.
The common sampling methods are compared in the \cref{fig:sample.pdf}(b). 
Spatial skipping based on the occupancy grid is the degraded form of our sampling method. Given the frequency value $f$ of the frequency grid traversed by ray $I$, the sampling frequency $f_{sample}$ should be ensured to be less than twice the detail frequency to comply with the sampling theorem, \ie the sampling frequency is $f_{sample} = 2f$.
Based on this formula, we can estimate an appropriate sampling interval according to the frequency of the content, eliminating the need to tune the sampling hyperparameters for different scenes.

\noindent\textbf{Training Loss:} The training loss is defined as
\begin{align}
   \mathcal{L}_{total}=&\mathcal{L}_{recon}(\hat{\mathbf{c}}, \mathbf{c}_{gt})+\\
   &\lambda_{dist}\mathcal{L}_{dist}(\mathbf{s_d},\mathbf{w})+\lambda_{depth} \mathcal{L}_{depth},
\end{align}
where $\mathcal{L}_{recon}(\hat{\mathbf{c}}, \mathbf{c}_{gt})=\sqrt{(\hat{\mathbf{c}} -\mathbf{c}_{gt}  )^2+\epsilon } $ is a color reconstruction loss, $\hat{\mathbf{c}}$ is the rendered pixel color, $\mathbf{c}_{gt}$ is the ground-truth pixel color, $\epsilon = 10^{-4}$. 
We regularize the density distribution in disparity through $\mathcal{L}_{dist}$, which is proposed by Mip-NeRF360~\cite{barron2022mip}, 
where $\mathbf{s_d}$ is the set of normalized ray distances and $\mathbf{w}$ is the set of weights. It penalizes the discreteness to encourage the formation of thinner surfaces. 
In contrast to Mip-NeRF 360 using a proposal network to obtain sampling suggestions, we compute this discrete version of sampling distribution regularization along the entire ray. 
$\mathcal{L}_{depth}$ is the depth loss between the estimated depth and the actual value from the sparse point cloud, which is used in early training to avoid incorrect geometry formation.
$\lambda_{dist}$, $\lambda_{depth}$ are the coefficients of loss. We provide more training details in the supplementary material.

\begin{table*}[]
\centering
\caption{\textbf{Quantitative comparisons on the \textsc{Multi-Frequency Dataset} dataset.} The \colorbox{red!40}{best}, \colorbox{orange!40}{second best}, and \colorbox{yellow!40}{third best} results are highlighted in red, orange, and yellow, respectively.}
    \resizebox{0.76\linewidth}{!}{
        \begin{tabular}{lccc|ccc}
        \hline
         & \multicolumn{3}{c|}{Structrure View (600$\times$ 600)} & \multicolumn{3}{c}{Detail View (4032$\times$ 3024)} \\ \hline
        Method|Metric & PSNR$^{\uparrow}$ & $\text{SSIM}^{\uparrow}$ & $\text{LPIPS}^{\downarrow}_{\text{(VGG)}}$ & $\text{PSNR}^{\uparrow}$ & $\text{SSIM}^{\uparrow}$ & $\text{LPIPS}^{\downarrow}_{\text{(VGG)}}$ \\ \hline
        TensoRF & 28.88 & 0.854 & 0.256 & 22.76 & 0.781 & 0.430 \\
        iNGP-Base & 30.27 & 0.893 & 0.216 & 23.63 & 0.784 & 0.408 \\
        iNGP-Big & \cellcolor{orange!40}30.97 & \cellcolor{orange!40}0.909 & \cellcolor{orange!40}0.183 & 24.00 & 0.786 & 0.398 \\
        Mip-NeRF360 & 30.79 & \cellcolor{yellow!40}0.906 & \cellcolor{yellow!40}0.188 & \cellcolor{yellow!40}24.16 & \cellcolor{yellow!40}0.792 & \cellcolor{orange!40}0.383 \\
        3D-GS & \cellcolor{yellow!40}30.85 & 0.897 & 0.191 & \cellcolor{orange!40}24.29 & \cellcolor{orange!40}0.802 & \cellcolor{yellow!40}0.390 \\  \hline
        Ours & \cellcolor{red!40}32.44 & \cellcolor{red!40}0.929 & \cellcolor{red!40}0.148 & \cellcolor{red!40}26.29 & \cellcolor{red!40}0.843 & \cellcolor{red!40}0.332 \\ \hline
        \end{tabular}   
    }

\label{tab:table1}
\end{table*}

\begin{figure*}[t!]
       \centering
       \includegraphics[width=0.9\linewidth]{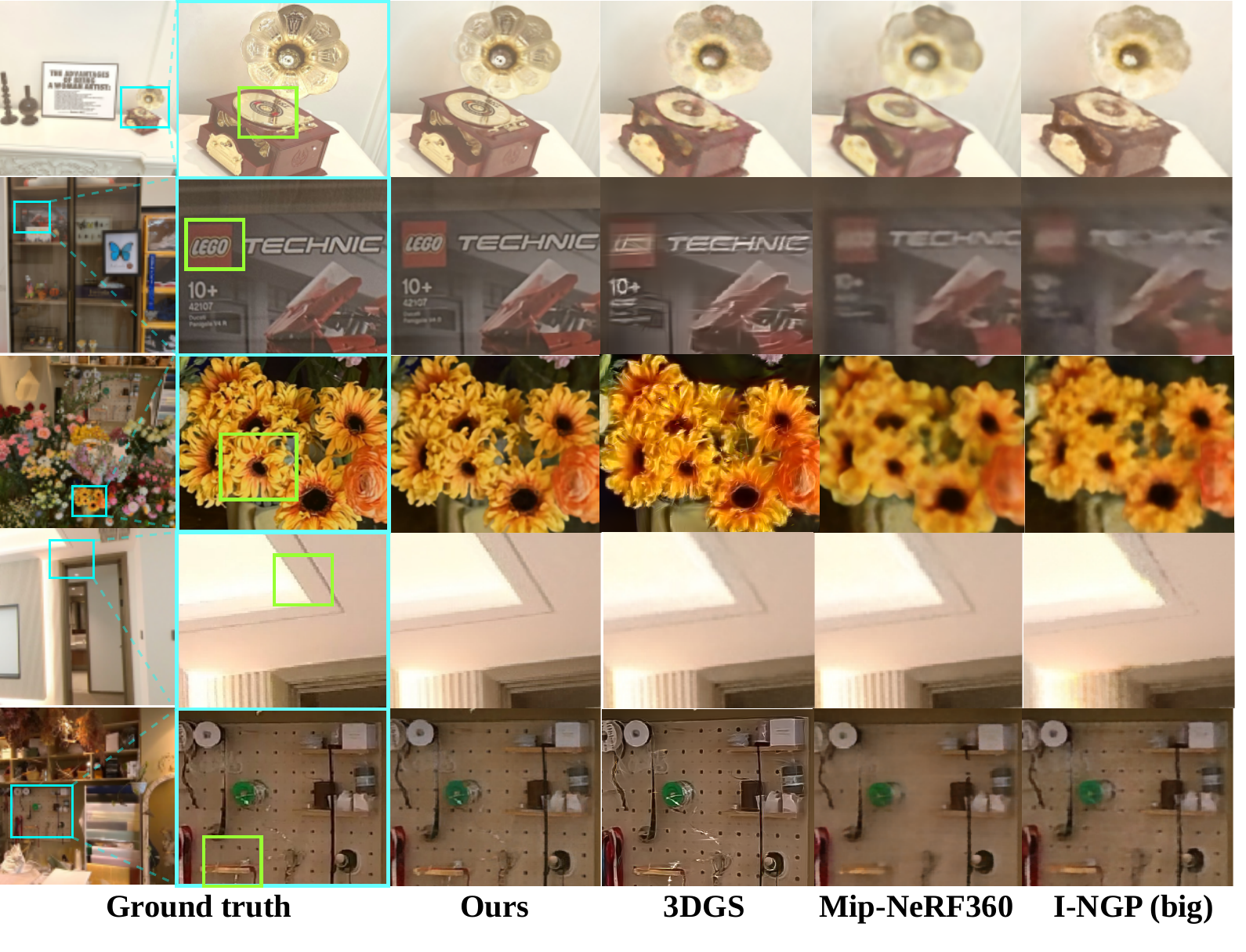}
       \caption{\textbf{Qualitative comparison.}We show the qualitative comparison results with 3D-GS~\cite{kerbl20233dgs}, Mip-NeRF360~\cite{barron2022mip}, and Instant-NGP~\cite{muller2022instant} on the \textsc{Multi-Frequency Dataset}.NeRFs tend to produce overly blurred results on high-frequency data, while 3D-GS often exhibits noticeable spiky artifacts. In contrast, our method faithfully preserves high-frequency details and geometric structures in scenes with large frequency variations.}
       \label{fig:figure2}
\end{figure*}

\section{Experiments}

\subsection{Experimental Settings}
\label{subsec:experimental settings}
\textbf{Datasets.} We use two kinds of datasets for evaluation. (1) Multi-frequency dataset with a large frequency span. The Multi-frequency dataset contains five indoor scenes, each of them contains dense normal-resolution (less than 2K) images for the entire structure of the scene and high-resolution (up to 4K) images focusing on the area of interest of the scene; (2) Standard dataset with a small frequency span.
The dataset contains five bounded scenes from the MipNeRF-360 dataset~\cite{barron2022mip} and two unbounded scenes from the Tanks\&Temples dataset~\cite{Knapitsch2017}. 
For all datasets, we use the standard settings that select one out of every eight images as the testing set, and the remaining images as the training set. PNSR, SSIM, and $\text{LPIPS}_{\text{(VGG)}}$ are employed as evaluation metrics.

\noindent\textbf{Baselines and Implementation.}
3D-GS~\cite{kerbl20233dgs}, MipNeRF-360~\cite{barron2022mip} and Instant-NGP~\cite{muller2022instant} and TensoRF~\cite{chen2022tensorf} are selected as our baselines . We set a constant batch size of $2^{18}$ for point samples.
Ray batch sizes vary based on the average number of sampled points per ray. 
We employ the Adam optimizer~\cite{kingma2014adam} for parameter training with $\beta_1=0.9, \beta_2=0.99, \epsilon=10^{-15}$. 
For the Multi-frequency dataset, we train all methods for 200k steps to achieve full convergence. For MipNeRF-360 and Tanks \& Temples dataset, our method and Instant-NGP are trained for 50k steps with the same batch size. For all datasets, we report results for a basic configuration of Instant-NGP (base) as well as a slightly larger hash table size (Big). All the methods are trained on a single RTX 4090 GPU.

\subsection{Results Analysis}
\label{subsec:Results}

In Tab.~\ref{tab:table1}, we report the quantitative evaluation results on the Multi-frequency dataset. Our method significantly surpasses all benchmark methods, both for normal-resolution images and high-resolution images. 
Our method does not only reconstruct the intact low-frequency structure of the scene, \eg, the clear ceiling in \textit{Home} and complete circuit board in \textit{Flower Shop}, but also recover sharp high-frequency contents, \eg, well-preserved of details of the phonograph in \textit{Music Room}, clear text of the lego in \textit{Home}, and vivid appearance of the flower in \textit{Flower Shop}.
Unlike the excessive blurring produced by NeRF-based methods, 3D-GS introduces numerous needle-like artifacts under large frequency spans. While it exhibits a sharp appearance, it fails to accurately capture the true details of the scene.
TensoRF\cite{chen2022tensorf} has the lowest PSNR metric due to its compression of high-frequency spatial information and is not included in the qualitative comparison. 
The results of Mip-NeRF 360~\cite{barron2022mip} are blurry in both low-frequency and high-frequency images because the continuity of MLP over-smooth significant frequency changes. 
The results of iNGP-Big appear sharper with more content, but due to the uniform partitioning of space and sampling, it lack clarity. 
This demonstrates that simply increasing capacity is insufficient to model multi-frequency content.
In contrast, our method can capture various frequencies within the scene through its frequency-aware design, thus enabling better restoration of both low-frequency structures and high-frequency details in multi-frequency contents.

\subsection{Ablation Studies}
\label{subsec:ablation studies}

\noindent\textbf{Frequency Issues in Standard Datasets.} We evaluate our method on MipNeRF-360~\cite{barron2022mip} and Tanks \& Temples~\cite{Knapitsch2017}, both object-centric datasets, and still demonstrates improvements. This indicates that multi-frequency issues exist in general datasets, as images with the same resolution can capture different frequency content. Since these datasets have a smaller frequency span than our Multi-frequency dataset, it does not significantly degrade rendering quality. Nevertheless, our multi-frequency optimization still benefits the high-fidelity modeling of various frequency components in the scene.

\noindent\textbf{Component Ablation.}We conduct ablation studies in the \textit{Music Room} of the Multi-frequency dataset. In the ablation studies, we individually deactivate the frequency-aware components, namely feature re-weighting, frequency-averaged sampling (FAS), and Adaptive Ray Marching (ARM), to test their effectiveness. Furthermore, we test the rendering results without our Frequency Grid by setting the frequency to a uniform value, thus disabling all the components mentioned above. 

\begin{table}[]
\centering
\caption{Quantitative comparisons on the MipNeRF-360~\cite{barron2022mip} dataset and Tanks\&Temples~\cite{Knapitsch2017} dataset.}
\vspace{-2pt}
\resizebox{0.99\linewidth}{!}{
\begin{tabular}{lccc|ccc}
\hline
Dataset & \multicolumn{3}{c|}{MipNeRF-360} & \multicolumn{3}{c}{Tanks\&Temples} \\ \hline
Method|Metric & $\text{PSNR}^{\uparrow}$ & $\text{SSIM}^{\uparrow}$ & $\text{LPIPS}^{\downarrow}_{\text{(VGG)}}$ & $\text{PSNR}^{\uparrow}$ & $\text{SSIM}^{\uparrow}$ & $\text{LPIPS}^{\downarrow}_{\text{(VGG)}}$ \\ \hline
TensoRF & 24.71 & 0.708 & 0.448 & 19.52 & 0.613 & 0.451 \\
iNGP-Base & 29.15 & 0.879 & 0.216 & 21.56 & 0.731 & 0.318 \\
iNGP-Big & 29.72 & 0.900 & 0.219 & 21.69 & 0.757 & 0.280 \\
Mip-NeRF360 & \cellcolor{red!40}31.49 & \cellcolor{yellow!40}0.916 & \cellcolor{yellow!40}0.179 & \cellcolor{yellow!40}22.22 & \cellcolor{yellow!40}0.759 & \cellcolor{yellow!40}0.257 \\
3D-GS & \cellcolor{yellow!40}30.95 & \cellcolor{orange!40}0.926 & \cellcolor{orange!40}0.167 & \cellcolor{orange!40}24.36 & \cellcolor{red!40}0.831 & \cellcolor{orange!40}0.213 \\
Ours & \cellcolor{orange!40}31.20 & \cellcolor{red!40}0.931 & \cellcolor{red!40}0.165 & \cellcolor{red!40}24.45 & \cellcolor{orange!40}0.821 & \cellcolor{red!40}0.205 \\ \hline
\end{tabular}
}
\label{tab:ablation}
\end{table}

\begin{table}[t]
\centering
\vspace{-0.8em}
\caption{Ablation Results for Different Components}
\vspace{-0.8em}
\setlength{\tabcolsep}{0.2mm}{
\resizebox{\linewidth}{!}{
\begin{tabular}{lccc|ccc}
\hline
\multicolumn{1}{c}{} & \multicolumn{3}{c|}{normal-res(600$\times$600)} & \multicolumn{3}{c}{high-res(4032$\times$ 3024)} \\ \hline 
\multicolumn{1}{c}{Setting} & PSNR$^{\uparrow}$ & SSIM$^{\uparrow}$ & LPIPS$^{\downarrow}_{\text{(VGG)}}$ & PSNR$^{\uparrow}$ & SSIM$^{\uparrow}$ & LPIPS$^{\downarrow}_{\text{(VGG)}}$ \\ \hline
A) w/o Frequency Grid & 31.95 & 0.930 & 0.181 & 24.90 & 0.910 & 0.316 \\
B) w/o Feature Re-weighting & \textbf{33.58} & \textbf{0.932} & 0.167 & 26.73 & 0.923 & 0.256 \\
C) w/o FAS & 33.50 & 0.929 & 0.163 & 25.84 & 0.921 & 0.268 \\
D) w/o adaptive RM & 32.30 & 0.928 & 0.165 & 25.42 & 0.922 & 0.255 \\
E) Our Complete Model & 33.52 & 0.931 & \textbf{0.162} & \textbf{26.97} & \textbf{0.924} & \textbf{0.250} \\ \hline
\end{tabular}
}
}
\vspace{-1.8em}
\label{tab:ablation}
\end{table}

As shown in Tab.~\ref{tab:ablation}, model $(\text{A})$ without the Frequency Grid performs worst, further confirming the effectiveness of our Frequency Grid in multi-frequency scenarios. Among all the frequency-aware components, disabling ARM causes the most significant performance drop. This is because high-frequency contents require denser sampling in the scene with a large frequency span, as described in Sec.~\ref{ssec:method_frequencyaware}.
Disabling frequency-averaged sampling reduces training frequency, hindering high-resolution rendering performance. Additionally, turning off feature re-weighting degrades high-resolution performance but improves normal-resolution results. This suggests that in multi-frequency scenarios, low-frequency signals overshadow high-frequency ones, causing performance drops, especially when feature space capacity is limited, such as with a smaller hash table.

\begin{figure}[!h]    
  \centering
  \vspace{-0.5em}
  \includegraphics[width=0.98\linewidth]{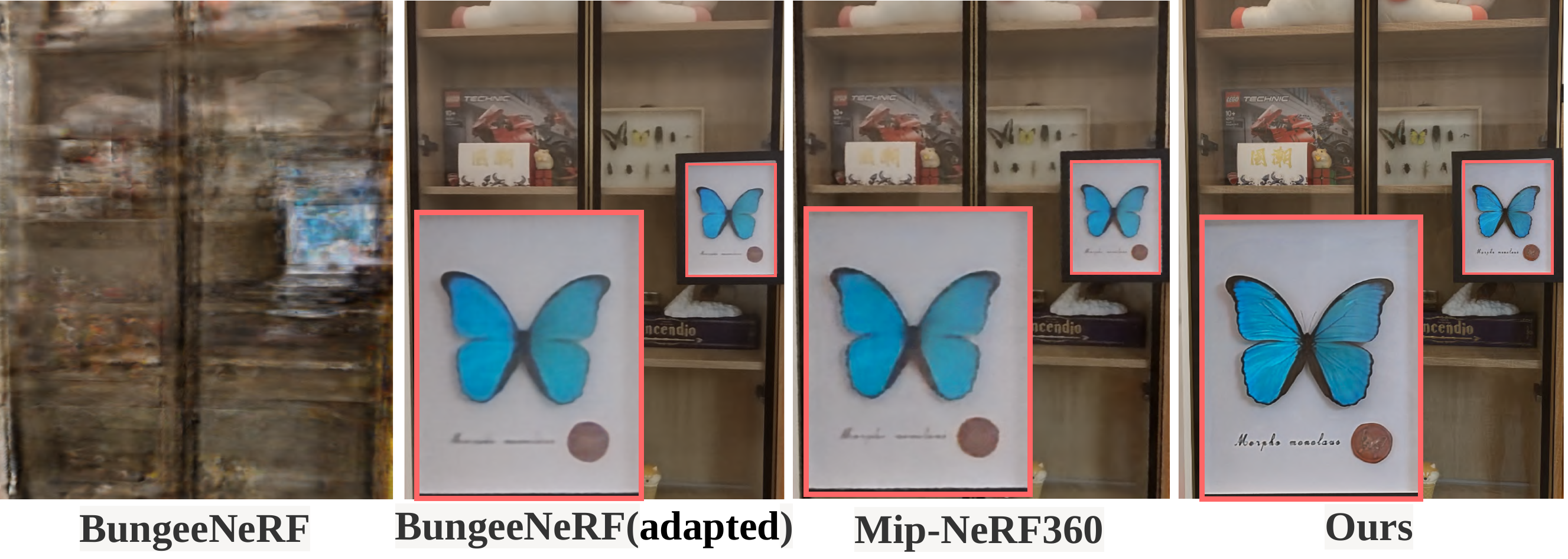}
   \vspace{-0.8em}
   \caption{Additional comparisons with BungeeNeRF}
   \vspace{-0.8em}
   \label{fig:Bugee}
\end{figure}

\noindent\textbf{Additional comparison with Multi-Scale Method.}
We also compared BungeeNeRF due to its ability in cross-scale rendering. This divide-and-conquer strategy is commonly used in Multi-Scale and Large-Scale tasks, but in real scenes, the frequency distribution isn’t strictly based on position. To make a fair comparison, we modified BungeeNeRF by assigning high-resolution cameras to the "close views" block and increasing the position encoding frequency to ($2^{11}$). Qualitative results in Fig.\ref{fig:Bugee} show that BungeeNeRF performs similarly to its baseline (Mip-NeRF) in multi-frequency scenarios. Quantitative results are in the supplementary.

\section{Conclusion}
\label{sec:relatedwork}
Our experiments demonstrate that we can effectively quantify the frequency components within a scene and perform targeted high-fidelity reconstruction. This is crucial for reconstructing scenes with tiny details. Our method outperforms all state-of-the-art approaches, and given the immersive effects, such scenes may inspire future research. While our grid-based representation enables 20 FPS rendering, there is still potential for further acceleration.
\newpage
{
    \small
    \bibliographystyle{ieeenat_fullname}
    \bibliography{main}
}
\clearpage
\setcounter{page}{1}
\maketitlesupplementary

\begin{figure*}[t]       
       \centering
       \includegraphics[width=0.8\linewidth]{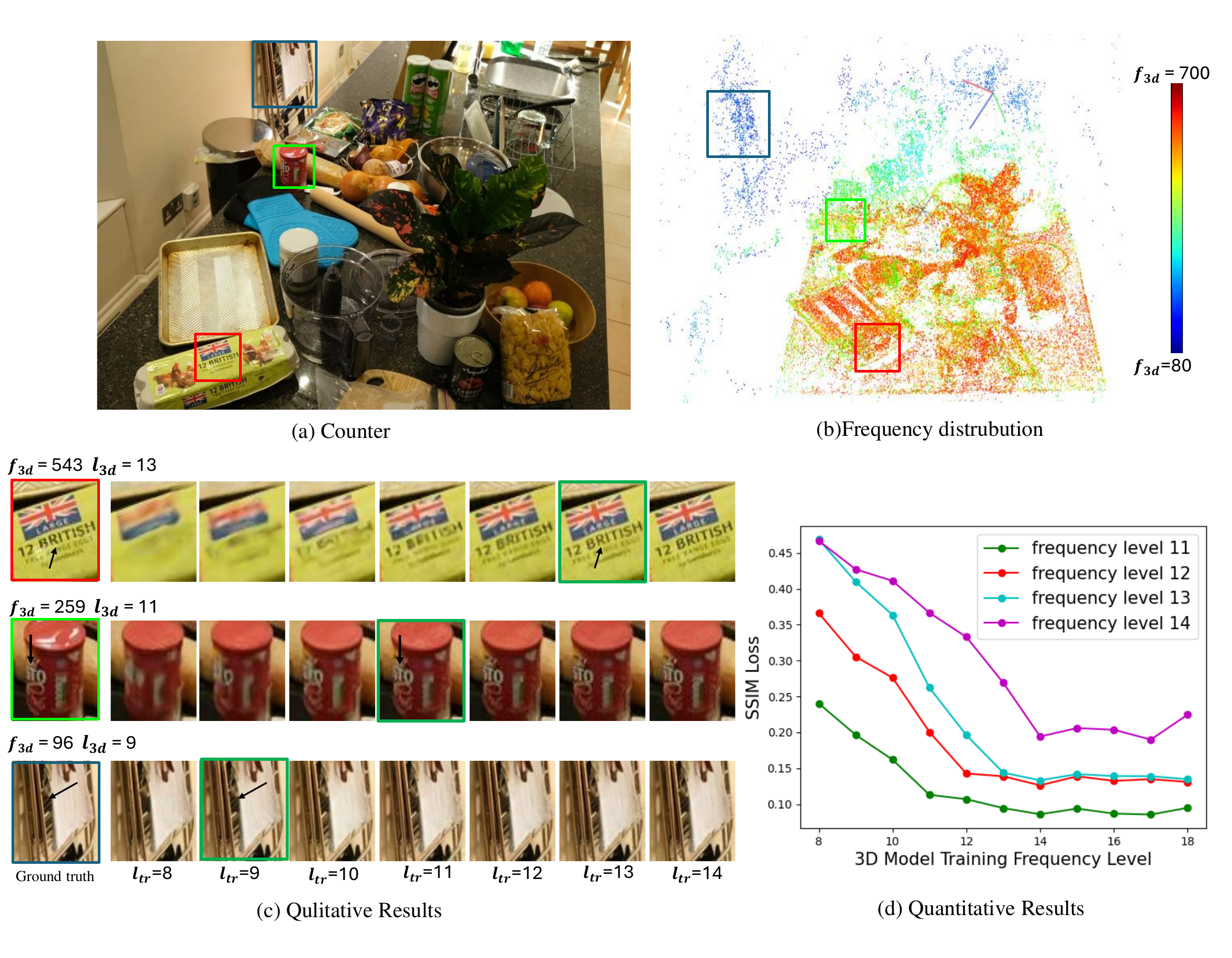}
           
           \caption{We test the effectiveness of frequency quantification on the real dataset NeRF-360-v2. (a) is a ground truth image from the \textit{counter} dataset. (b) is the colored point cloud after the 3D frequency initialization of all 3D points, where colors leaning towards blue indicate lower frequencies and those towards red indicate higher frequencies. It can be observed that there is a trend of higher frequencies with smaller depths. Moreover, different regions at the same depth also exhibit varying distributions of 3D frequencies. In (a), we selected three patches with different frequency levels $\ell_{3D}$ 9, 11, and 13, represented by blue, green, and red, respectively. (c) is the rendering results of these three patches at training frequency level $\ell_{tr}$ ranging from 8 to 14. It can be noticed that details are not well recovered when $\ell_{tr}$ is lower than the quantified $\ell_{3D}$, and when $\ell_{tr}$ exceeds $\ell_{3D}$, there is no significant improvement in the quality of the rendering images. (d) shows the distribution of average SSIM Loss for patches of different frequency levels in the \textit{counter} dataset at various training frequency levels $\ell_{tr}$. Once $\ell_{tr}$ reaches the quantified $\ell_{3D}$, there is no significant decrease in loss.}
           \label{fig:freq}
           \vspace{-0.5em}
\end{figure*}

This supplementary material includes video results for scenes from the Multi-frequency datasets. In the following sections, we first introduce additional implementation details (Sec.~\ref{sec:add_details}). Moreover, we provide additional experiments on frequency level evaluation from 2D to 3D on the real dataset Mip-NeRF360-v2 (Sec.~\ref{sec:freq}). Finally, we present more experimental results (Sec.~\ref{sec:add_exp}).

\section{Additional Implementation Details}
\label{sec:add_details}

\subsection{Dataset}
We captured our dataset using two different cameras. We collected normal-resolution images with a panoramic camera to achieve a more comprehensive 360-degree field of view for better scene structure reconstruction. And high-resolution images were captured with a DSLR (Digital Single-Lens Reflex) camera. To obtain camera poses, we perform Structure from Motion reconstruction for both the panoramic and high-resolution images. We use shared intrinsic between all images of the same camera model in a scene, and calibrate using the OpenCV radial distortion model. We then project the panoramic images into six 600$\times$600 perspective images, each with FOV of 60$^\circ$ to accommodate the perspective camera model commonly used in most NeRF models. We adopt a common dataset splitting method, selecting one out of every eight panoramic/high-resolution images as the test set, with the remainder constituting the training set.

\subsection{Architecture details}
We adopt a setup similar to Instant-NGP~\cite{muller2022instant}, utilizing 16 grid scales with the maximum resolution being 2048$\times$scene size and the minimum resolution being 16, employing 2 feature channels per level. In our dataset, due to the larger scene sizes, we set the size of the hash table storing feature vector for each level to $2^{23}$ to mitigate the impact of hash collisions on scene representation. For other general scenes, we use an identical hash table size of $2^{19}$ to Instant-NGP~\cite{muller2022instant}. The fetched hash feature vectors are down-weighted before being concatenated and fed to a one-layer MLP with 64 hidden units to get the scene features and the volume densities. Subsequently, the scene features are concatenated with the spherical harmonics encoding of the view directions, which is then input to a subsequent two-layer MLP of width 64 to yield the RGB colors.

\subsection{Frequency Grid}
To represent the frequency distribution in the 3D space, we maintain a frequency grid with a resolution of 128$\times$AABB, where AABB, short for Axis-Aligned Bounding Box, denotes the scene size. For each scene, we adjust the AABB based on the 3D points from the SfM reconstruction to ensure it encompasses the majority of the 3D points. Each grid cell stores the frequency level as a uint8 number. 

\noindent{\textbf{Initialization.}} Once we have the 2D frequencies of all training patches, we first calculate the 3D frequency of each 3D point $p_i$. After that, each 3D point is reprojected to obtain a set of observation patches $\{P_{ij}|j=1,...,n\}$ and derive a set of 3D frequencies $\{f_{{3D}_{ij}}|j=1,...,n\}$ with the depth of the point. To mitigate the influence of noisy patches, we take the median of this set as the 3D frequency $f_{{3D}_i}$ for that point. Assuming that the frequencies at each level are $\{f_{{3D}_{\ell}}|\ell=0,...,n_{\ell}\}$, we take the frequency level $\ell_i$ as $\underset{\ell}{\arg\min}(|f_{{3D}_\ell}-f_{{3D}_i}|)$. The frequency grid is then initialized to the maximum of the frequency levels of all 3D points within the grid.

\noindent{\textbf{Re-weighting.}} Unlike Instant-NGP~\cite{muller2022instant}, which directly concatenates feature vectors as the input for the tiny MLP, we take into account the 3D frequency at that point and re-weight different frequency components accordingly. Instead, we use the quantified frequency level $\ell$ as a threshold and apply a down-weighting to frequency components that are higher than $\ell$.
We compute the down-weighting factor $w$ using an approximation for $\text{erf}(x)$:
\begin{align}
    \text{erf}(x)\approx \text{sign}(x)\sqrt{1-\text{exp}(-(4/\pi)x^2)}
\end{align}

\noindent{\textbf{Updating.}} We update the grids after every 1024 training iterations by the following steps. We first render the depth of the center pixel of a training patch $P_i$. Then, the 2D frequency of the patch is projected to the corresponding 3D point to obtain its 3D frequency $f_{{3D}_i}$ and frequency level $\ell_i$. Finally, the value $\ell$ of the frequency grid where the 3D point resides is then updated to $\max(\ell_i, \ell)$.

\noindent{\textbf{Frequency-averaged sampling(FAS).}} We divides the training batch into $N$  segments based on the frequency quantization results. The sampling frequency is evenly distributed within a preset range of [1, 3], meaning that the highest frequency content is sampled with a probability three times that of the lowest frequency. In our experiments, we found that this is a more stable setting compared to directly using the frequency ratio as the sampling proportion.

\subsection{Loss Functions}
As described in the main paper, the training loss is defined as
\begin{equation}
\begin{aligned}
   \mathcal{L}_{total}=&\mathcal{L}_{recon}(\hat{\mathbf{c}}, \mathbf{c}_{gt})+\lambda_{depth} \mathcal{L}_{depth}(\hat{\mathbf{d}},\mathbf{d}_{gt})+ \\
   &\lambda_{dist}\mathcal{L}_{dist}(\mathbf{s_d},\mathbf{w}),
\end{aligned}
\end{equation}
where the first term $\mathcal{L}_{recon}(\hat{\mathbf{c}}, \mathbf{c}_{gt})=\sqrt{(\hat{\mathbf{c}} -\mathbf{c}_{gt}  )^2+\epsilon } $ is a color reconstruction loss~\cite{barron2022mip}, $\hat{\mathbf{c}}$ is the rendered pixel color, $\mathbf{c}_{gt}$ is the ground-truth pixel color, and $\epsilon = 10^{-4}$, and the last term is the regularization loss.

The depth loss $\mathcal{L}_{depth}$ of the sampled ray is defined by
\begin{align}
    \mathcal{L}_{depth}(\hat{\mathbf{d}}, \mathbf{d}_{gt})=\sqrt{(\hat{\mathbf{d}} -\mathbf{d}_{gt}  )^2+\epsilon }
\end{align}
where the depth of a ray is computed by the weighted sum of the sampled distance that $d=\sum_i w_i t_i$, and $\{w_i\}$ are the weights computed by the volume rendering. We only use the depth loss in early training for pixels with GT depth from the sparse point cloud to avoid incorrect geometry structure.

The regularization loss is proposed by Mip-NeRF360~\cite{barron2022mip}. We use it to prevent floaters and background collapse, which is defined as

\begin{equation}
\begin{aligned}
    \mathcal{L}_{\text{dist}}(\mathbf{s_d}, \mathbf{w}) & =\sum_{i, j} w_{i} w_{j}\left|\frac{s_{i}+s_{i+1}}{2}\frac{s_{j}+s_{j+1}}{2}\right|+ \\
    &\frac{1}{3} \sum_{i} w_{i}^{2}\left(s_{i+1}-s_{i}\right),
\end{aligned}
\end{equation}

where $\mathbf{s_d}$ is the set of normalized ray distances and $\mathbf{w}$ is the set of weights. It penalizes the discreteness to encourage the formation of thinner surfaces. In contrast to Mip-NeRF360 using a proposal network to obtain sampling suggestions, we compute this discrete version of sampling distribution regularization along the entire ray. 

The hyperparameters $\lambda_{dist}$, $\lambda_{depth}$ are used to balance the data terms and the regularize; we set $\lambda_{dist}=0.01, \lambda_{depth}=0.001$ for all experiments.

\begin{figure}[!h]       
       \centering
       \includegraphics[width=0.98\linewidth]{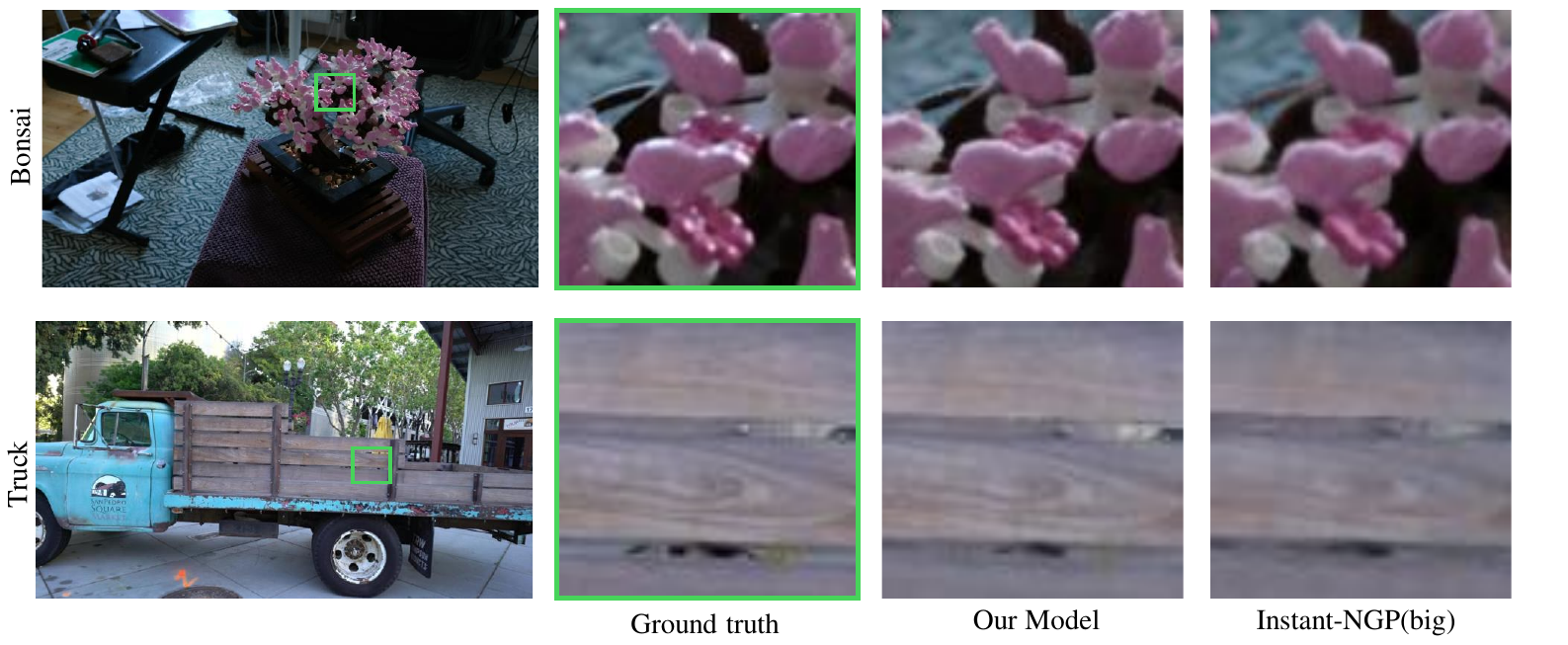}
           
           \caption{Qualitative comparisons with the Instant-NGP~\cite{muller2022instant} that has a larger hash table size (Big) on the Mip-NeRF360-v2~\cite{barron2022mip} and Tank and Temples~\cite{Knapitsch2017} dataset.}
           \label{fig:standard_res}
           \vspace{-0.5em}
\end{figure}

\section{Evaluate Frequency Level from 2D to 3D}
\label{sec:freq}
In this section, we further demonstrate the effectiveness of frequency quantification from 2D to 3D using the real dataset Mip-NeRF360-v2.

\noindent\textbf{Visualization of Frequency Distribution.} As described in the main paper, we reproject each 3D point from the sparse point cloud back into all the observation images. Then we calculate the 3D frequency set $S$ based on all the corresponding patches. The median of $S$ is taken as the 3D frequency for that point. Fig.~\ref{fig:freq}(b) shows a visualization of the 3D frequency distribution of all 3D points after initialization for the dataset \textit{counter} in Mip-NeRF360-v2, where the color of the points indicates the corresponding 3D frequency, with points closer to blue indicating a lower frequency and those closer to red indicating a higher frequency. Fig.~\ref{fig:freq}(a) represents the ground truth image, where the \textcolor{blue}{blue}, \textcolor{green}{green}, and \textcolor{red}{red} boxes represent three patches with 3D frequencies from low to high as shown in Fig.~\ref{fig:freq}(b).

\begin{figure*}[h]       
       \centering
       \includegraphics[width=0.9\linewidth]{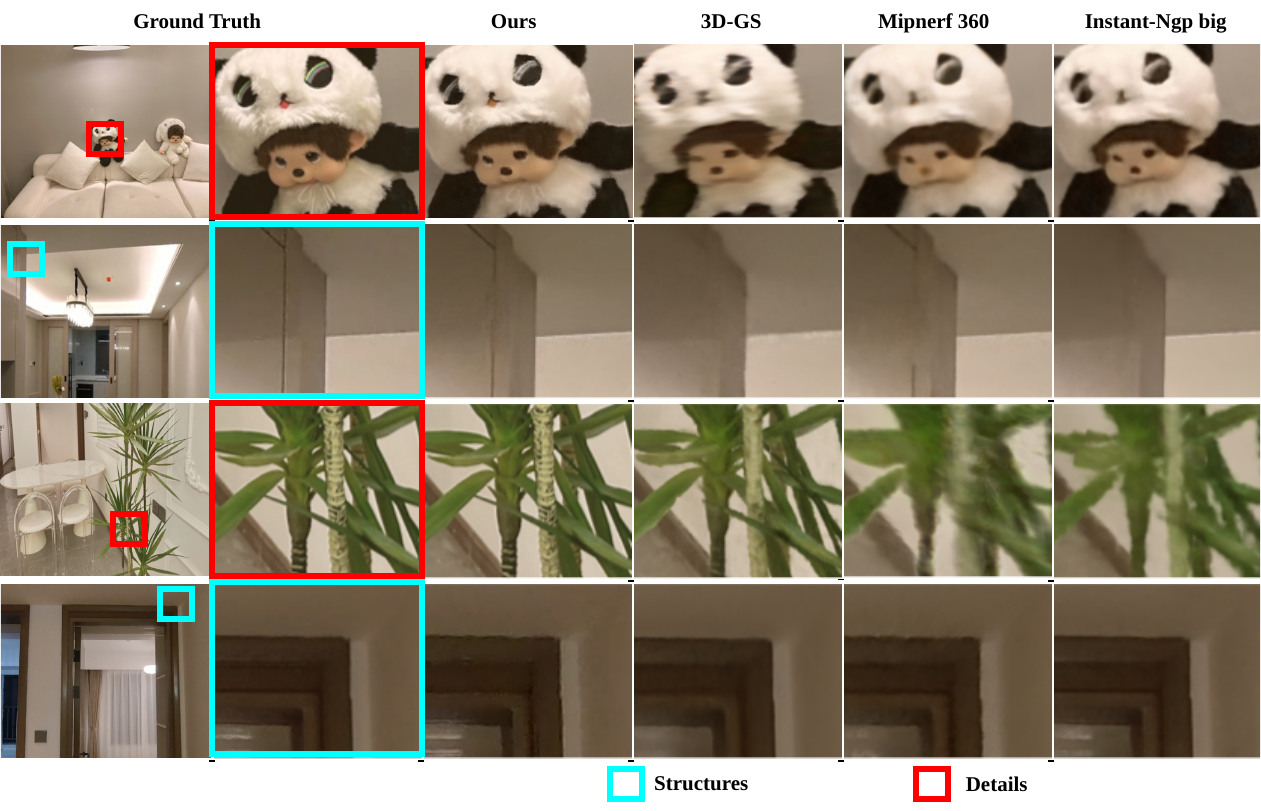}
           
           \caption{Additional visual comparisons on the Multi-frequency dataset.}
           \label{fig:result}
           \vspace{-0.5em}
\end{figure*}

\begin{table}[t]
\centering
\tabcolsep=0.2cm
\resizebox{1.0\linewidth}{!}{
\begin{tabular}{lccccc}
\hline
Method & room & counter & kitchen & bonsai & average\\ \hline
TensoRF & 0.791	& 0.697	& 0.560	& 0.783	& 0.708 \\
INGP-Base & 0.893 & 0.845 & 0.857 & 0.924 & 0.879 \\
INGP-Big & 0.900 & 0.868 & 0.907 & 0.922 & 0.900 \\ 
Mip-NeRF360 & 0.913& 0.895& 0.920& 0.939 & 0.916 \\
3D-GS & 0.914& 0.905& 0.922& 0.938 & 0.919 \\
Ours & 0.936 & 0.908 & 0.931 & 0.946 & \textbf{0.931} \\ \hline
\end{tabular}
}
\caption{SSIM on the Mip-NeRF360-v2 dataset} 
\label{tab:table8}
\end{table}

\begin{table}[t]
\centering
\tabcolsep=0.2cm
\resizebox{1.0\linewidth}{!}{
\begin{tabular}{lccccc}
\hline
Method & room & counter & kitchen & bonsai & average\\ \hline
TensoRF & 0.419	&0.469&	0.516&	0.389&	0.448 \\
INGP-Base & 0.242& 0.255& 0.170& 0.198 & 0.216 \\
INGP-Big & 0.254& 0.256& 0.158&	0.209&	0.219 \\ 
Mip-NeRF360 & 0.211& 0.203& 0.126& 0.177 &0.179 \\
3D-GS & 0.220 & 0.204 & 0.129 & 0.205 & 0.189 \\
Ours & 0.191 & 0.184 & 0.123 & 0.159 & \textbf{0.165} \\ \hline
\end{tabular}
}
\caption{LPIPS on the Mip-NeRF360-v2 dataset} 
\label{tab:table9}
\end{table}
\begin{table}[t]
\centering
\tabcolsep=0.2cm
\resizebox{1.0\linewidth}{!}{
\begin{tabular}{lccccc}
\hline
Method & room & counter & kitchen & bonsai & average\\ \hline
TensoRF & 26.88 & 23.39 & 23.12 & 25.46 & 24.71 \\
INGP-Base & 30.31 & 26.21 & 29.00 & 31.08 & 29.15 \\
INGP-Big & 30.19 & 27.27 & 30.86 & 30.57 & 29.72 \\ 
Mip-NeRF360 & 31.40 & 29.44 & 32.02 & 33.11 & \textbf{31.49} \\
3D-GS & 30.63 & 28.70 & 30.32 & 31.98 & 30.95 \\
Ours & 31.45 & 29.19 & 31.41 & 32.75 & 31.20 \\ \hline
\end{tabular}
}
\caption{PSNR on the Mip-NeRF360-v2 dataset} 
\label{tab:table7}
\end{table}

\noindent\textbf{Qualitative Results.} Fig.~\ref{fig:freq}(c) depicts the visual comparison of the rendering results under varying training frequencies of the three patches mentioned above, where the boxed patches represent the rendering results under the quantified 3D frequency level $\ell_{3D}$. It is clearly demonstrated that when the training frequency level is lower than $\ell_{3D}$, the network is unable to fully recover the detailed information. Conversely, when the training frequency exceeds the quantified 3D frequency, the network does not yield better results either.

\noindent\textbf{Quantitative Results.} Furthermore, in Fig.\ref{fig:freq}(d), the lines in green, red, blue, and purple correspond to patches with 3D frequency levels of 11, 12, 13, and 14, respectively. With the escalation of the training frequency from 8 to 14, there is a progressive reduction in the SSIM loss for the generated patches. Upon reaching the quantified 3D frequency for each patch with the training frequency, the SSIM loss reduction becomes more consistent. This observation suggests two key points: firstly, the necessary minimum NeRF frequency level for the complete reconstruction of the scene's diverse 3D frequency structures and textures is variable; secondly, the 3D frequency estimation we employ for the content provides an accurate reflection of their actual frequencies.

\section{More Experimental Details}
\label{sec:add_exp}


\begin{table*}[!t]
\centering
\tabcolsep=0.1cm
\resizebox{0.8\linewidth}{!}{%
\begin{tabular}{lcccccl}
\hline
Setting & FlowerShop & Home & DollsRoom & MusicRoom & PlantRoom & Average \\ \hline
\textbf{normal-res(600$\times$600)} & \multicolumn{1}{l}{} & \multicolumn{1}{l}{} & \multicolumn{1}{l}{} & \multicolumn{1}{l}{} & \multicolumn{1}{l}{} &  \\ \hline
w/o Feature Re-weighting & 28.19 & 32.93 & 34.12 & 33.46 & 33.42 & 32.42 \\
w/o FAS & 26.36 & 32.93 & 34.18 & 33.50 & 33.59 & 32.11 \\
w/o Interval Adjustment & 27.12 & 32.62 & 34.09 & 32.30 & 32.96 & 31.82 \\
Our Complete Model & 28.23 & 32.91 & 34.20 & 33.52 & 33.36 & 32.44 \\ \hline
\textbf{high-res(4032$\times$3024)} & \multicolumn{1}{l}{} & \multicolumn{1}{l}{} & \multicolumn{1}{l}{} & \multicolumn{1}{l}{} & \multicolumn{1}{l}{} &  \\ \hline
w/o Feature Re-weighting & 24.62 & 26.14 & 28.41 & 26.54 & 24.55 & 26.05 \\
w/o FAS & 24.01 & 26.47 & 28.74 & 26.84 & 23.98 & 25.97 \\
w/o Interval Adjustment & 23.82 & 25.79 & 28.31 & 26.02 & 23.81 & 25.55 \\
Our Complete Model & 24.86 & 26.24 & 28.75 & 26.97 & 24.63 & 26.29 \\ \hline
\end{tabular}
}
\caption{Ablation Studies on Multi-frequency dataset}
\label{tab:ablation1}
\vspace{-1.2em}
\end{table*}

\subsection{Quantitative Results on Standard Datasets}
We compare our methods against our baselines on the standard datasets whose scenes have a smaller frequency span and size.
The quantitative results are shown in the Tab.~\textcolor{red}{2}.
Here we show the qualitative comparisons with the Instant-NGP~\cite{muller2022instant}, as depicted in Fig.~\ref{fig:standard_res}.
Our methods render sharper and clearer high-frequency contents than the Instant-NGP, Indicating that while our frequency-aware framework is designed to handle high-quality model scene structures and details in scenarios with significant frequency disparities, it still generalizes well on standard datasets, enhancing rendering quality, particularly in high-frequency details.

\subsection{More Ablation Studies}
\begin{figure*}[!h]       
       \centering
       \includegraphics[width=0.98\linewidth]{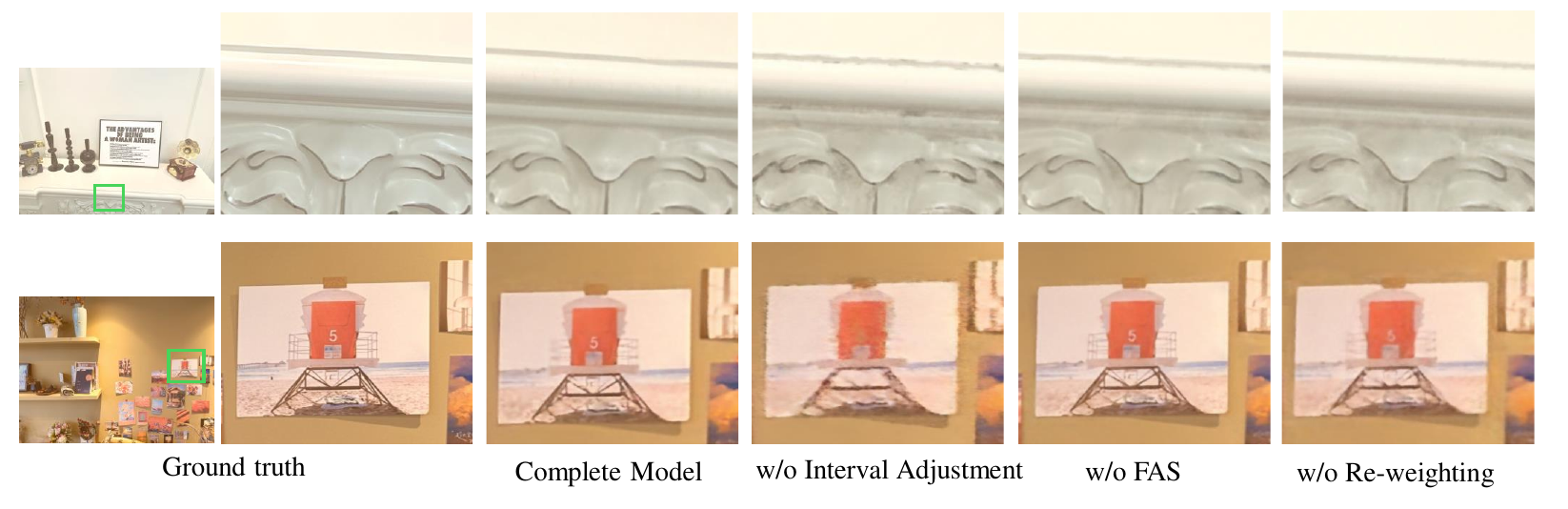}
           \vspace{-1.2em} 
           \caption{Additional visual comparisons with different settings of our method.}
           \label{fig:ablation}
           \vspace{0em}
\end{figure*}

\noindent\textbf{Component Ablation.} We conducted ablation experiments in each scene, and the results are shown in Tab.~\ref{tab:ablation1}. The results indicate that the impact of different features on overall performance varies across scenes of different scales.
In particular, in high-frequency scenes captured at close range, such as the "Flower Shop", the sampling interval adjustment has a more pronounced effect. This tendency is particularly evident in large-scale datasets or close-range captures. 
As shown in Figure.~\ref{fig:ablation}, close-range high-frequency content becomes blurred in the absence of sampling interval adjustment, which aligns with the description in Section 4.2 of the paper. Due to variations in the proportion of high-frequency data within scenes, the efficacy of FAS also varies. Balancing training batches sometimes enhances high-frequency effects, while at other times it may diminish them, depending on the distribution of scene data. Feature re-weighting enhances the network's efficiency in utilizing various frequency ranges, particularly when there is abundant scene content and limited network capacity.

\subsection{Per-Scene Metrics}
We provide the per-scene results on the Multi-frequency dataset, Tanks\&Temples dataset, and Mip-NeRF360-v2 dataset in Tab.~\ref{tab:table1} -~\ref{tab:table9}. The results are reported in the metrics of PSNR, SSIM, and LPIPS.
We provide more visual comparisons on the Multi-frequency dataset in Fig.~\ref{fig:result}.

\begin{table*}[!h]
\centering
\tabcolsep=0.1cm

\caption{PSNR on the Multi-frequency dataset(structure view)}
\vspace{-0.8em} 
\resizebox{0.8\linewidth}{!}{%
\begin{tabular}{lcccccc}
\hline
\textbf{PSNR} & FlowerShop & Home & DollsRoom & MusicRoom & PlantRoom & Average\\ \hline
TensoRF & 24.36 & 29.63 & 30.78 & 30.69 & 28.94 & 28.88 \\ 
INGP-Base & 25.31 & 30.10 & 32.07 & 31.78 & 32.07 & 30.27 \\ 
INGP-Big & 26.22 & 30.94 & 32.75 & 32.25 & 32.67 & 30.97 \\ 
Mip-NeRF360 & 26.90 & 32.01 & 31.28 & 32.59 & 31.15 & 30.79 \\
3D-GS & 27.02 & 32.08 & 31.30 & 32.74 & 31.17 & 30.85 \\ 
Ours & \textbf{28.23} & \textbf{32.91} & \textbf{34.2} & \textbf{33.52} & \textbf{33.36} & \textbf{32.44} \\ \hline 
\end{tabular}
}

\label{tab:table1}
\end{table*}

\begin{table*}[!h]
\centering
\tabcolsep=0.1cm
\caption{SSIM on the Multi-frequency dataset(structure view)}
\vspace{-0.8em} 
\resizebox{0.8\linewidth}{!}{%
\begin{tabular}{lcccccc}
\hline
\textbf{SSIM} & FlowerShop & Home & DollsRoom & MusicRoom & PlantRoom & Average\\ \hline
TensoRF & 0.740 & 0.884 & 0.893 & 0.890 & 0.862 & 0.854 \\ 
INGP-Base & 0.791 & 0.902 & 0.925 & 0.921 & 0.924 & 0.893 \\ 
INGP-Big & 0.830 & 0.918 & 0.934 & 0.930 & 0.933 & 0.909 \\ 
Mip-NeRF360 & 0.851 & 0.935 & 0.903 & 0.923 & 0.896 & 0.906 \\
3D-GS & 0.846 & 0.931 & 0.900 & 0.910 & 0.899 & 0.897 \\ 
Ours & \textbf{0.890} & \textbf{0.942} & \textbf{0.946} & \textbf{0.931} & \textbf{0.937} & \textbf{0.929} \\ \hline 
\end{tabular}
}

\label{tab:table2}
\end{table*}

\begin{table*}[!h]
\centering
\tabcolsep=0.1cm
\caption{LPIPS on the Multi-frequency dataset(structure view)}
\vspace{-0.8em} 
\resizebox{0.8\linewidth}{!}{%
\begin{tabular}{lcccccc}
\hline
\textbf{LPIPS} & FlowerShop & Home & DollsRoom & MusicRoom & PlantRoom & Average\\ \hline
TensoRF & 0.310 & 0.230 & 0.233 & 0.240 & 0.266 & 0.256 \\ 
INGP-Base & 0.267 & 0.205 & 0.206 & 0.207 & 0.196 & 0.216 \\ 
INGP-Big & 0.210 & 0.169 & 0.181 & 0.184 & 0.172 & 0.183 \\ 
Mip-NeRF360 & 0.181 & 0.158 & 0.208 & 0.187 & 0.206 & 0.188 \\
3D-GS & 0.177 & 0.164 & 0.217 & 0.189 & 0.211 & 0.191 \\ 
Ours & \textbf{0.148} & \textbf{0.130} & \textbf{0.146} & \textbf{0.162} & \textbf{0.152} & \textbf{0.148} \\ \hline 
\end{tabular}
}

\label{tab:table3}
\end{table*}


\begin{table*}[!h]
\centering
\tabcolsep=0.1cm
\caption{PSNR on the Multi-frequency dataset(detail view)}
\vspace{-0.8em} 
\resizebox{0.8\linewidth}{!}{
\begin{tabular}{lcccccc}
\hline
\textbf{PSNR} & FlowerShop & Home & DollsRoom & MusicRoom & PlantRoom & Average\\ \hline
TensoRF & 23.23 & 16.54 & 26.93 & 25.19 & 21.91 & 22.76 \\ 
INGP-Base & 22.27 & 22.82 & 26.75 & 24.49 & 21.83 & 23.63 \\ 
INGP-Big & 22.65 & 23.36 & 27.02 & 24.90 & 22.06 & 24.00 \\ 
Mip-NeRF360 & 23.27 & 24.28 & 25.98 & 25.28 & 21.97 & 24.16 \\
3D-GS & 23.42 & 24.46 & 26.11 & 25.41 & 22.01 & 24.29 \\ 
Ours & \textbf{24.86} & \textbf{26.24} & \textbf{28.75} & \textbf{26.97} & \textbf{24.64} & \textbf{26.29} \\ \hline 
\end{tabular}
}

\label{tab:table4}
\end{table*}

\begin{table*}[!h]
\centering
\tabcolsep=0.1cm
\caption{SSIM on the Multi-frequency dataset(detail view)}
\vspace{-0.8em} 
\resizebox{0.8\linewidth}{!}{
\begin{tabular}{lcccccc}
\hline
\textbf{SSIM} & FlowerShop & Home & DollsRoom & MusicRoom & PlantRoom & Average\\ \hline
TensoRF & 0.726 & 0.655 & 0.791 & 0.882 & 0.852 & 0.781 \\ 
INGP-Base & 0.716 & 0.717 & 0.769 & 0.868 & 0.848 & 0.784 \\ 
INGP-Big & 0.720 & 0.722 & 0.770 & 0.870 & 0.849 & 0.786 \\ 
Mip-NeRF360 & 0.686 & 0.731 & 0.795 & 0.884 & 0.863 & 0.792 \\
3D-GS & 0.735 & 0.758 & 0.791 & 0.893 & 0.833 & 0.802 \\ 
Ours & \textbf{0.770} & \textbf{0.813} & \textbf{0.817} & \textbf{0.924} & \textbf{0.892} & \textbf{0.843} \\ \hline 
\end{tabular}
}

\label{tab:table5}
\end{table*}

\begin{table*}[!h]
\centering
\tabcolsep=0.2cm
\vspace{-0.8em} 
\caption{LPIPS on the Multi-frequency dataset(detail view)}
\vspace{-0.8em} 
\resizebox{0.8\linewidth}{!}{
\begin{tabular}{lcccccc}
\hline
\textbf{LPIPS} & FlowerShop & Home & DollsRoom & MusicRoom & PlantRoom & Average\\ \hline
TensoRF & 0.459 & 0.592 & 0.415 & 0.316 & 0.367 & 0.430 \\ 
INGP-Base & 0.466 & 0.500 & 0.404 & 0.323 & 0.346 & 0.408 \\ 
INGP-Big & 0.485 & 0.449 & 0.401 & 0.316 & 0.337 & 0.398 \\ 
Mip-NeRF360 & 0.421 & 0.459 & 0.413 & 0.292 & 0.343 & 0.383 \\
3D-GS & 0.422 & 0.454 & 0.423 & 0.306 & 0.347 & 0.390 \\ 
Ours & \textbf{0.384} & \textbf{0.361} & \textbf{0.367} & \textbf{0.250} & \textbf{0.302} & \textbf{0.332} \\ \hline 
\end{tabular}
}

\label{tab:table6}
\end{table*}


\begin{table*}[!h]
\centering
\renewcommand\arraystretch{1.1}
\vspace{-0.8em} 
\caption{Results on the Tanks\&Temples dataset}
\vspace{-0.8em} 
\resizebox{0.9\linewidth}{!}{
\begin{tabular}{l|ccc|ccc|ccc}
\hline
Dataset & \multicolumn{3}{c|}{Train} & \multicolumn{3}{c|}{Truck} & \multicolumn{3}{c}{Average} \\ \hline
Method|Metric & PSNR$^{\uparrow}$ & SSIM$^{\uparrow}$ & LPIPS$^{\downarrow}_{\text{(VGG)}}$ & PSNR$^{\uparrow}$ & SSIM$^{\uparrow}$ & LPIPS$^{\downarrow}_{\text{(VGG)}}$ & PSNR$^{\uparrow}$ & SSIM$^{\uparrow}$ & LPIPS$^{\downarrow}_{\text{(VGG)}}$ \\ \hline
TensoRF & 18.73 & 0.569 & 0.490 & 20.30 & 0.657 & 0.411 & 19.52 & 0.613 & 0.451 \\
INGP-Base & 20.43 & 0.684 & 0.367 & 22.69 & 0.777 & 0.269 & 21.56 & 0.731 & 0.318 \\
INGP-Big & 20.39 & 0.711 & 0.332 & 22.98 & 0.803 & 0.227 & 21.69 & 0.757 & 0.280 \\
Mip-NeRF360 & 19.52 & 0.660 & 0.354 & 24.91 & 0.857 & 0.159 & 22.22 & 0.759 & 0.257 \\
3D-GS & 23.06 & 0.813 & 0.200  & 25.66 & 0.849 & 0.226  & 24.36 & 0.831 & 0.213 \\
Ours & 23.13 & 0.802 & 0.197 & 25.77 & 0.841 & 0.210 & 24.45 & 0.821 & 0.205 \\ \hline
\end{tabular}
}

\label{tab:table1}
\end{table*}

\begin{table*}[!h]
\footnotesize
\centering
\tabcolsep=0.01cm
\vspace{-0.8em} 
\caption{Quantitative comparisons}
\vspace{-0.8em} 
\resizebox{0.8\linewidth}{!}{%
\begin{tabular}{lccc|ccc}
\hline
\multicolumn{1}{c}{} & \multicolumn{3}{c|}{Structrure View(600$\times$600)} & \multicolumn{3}{c}{Detail View(4032$\times$ 3024)} \\ \hline 
\multicolumn{1}{c}{Method(Mem)} & PSNR$^{\uparrow}$ & SSIM$^{\uparrow}$ & LPIPS$^{\downarrow}$ & PSNR$^{\uparrow}$ & SSIM$^{\uparrow}$ & LPIPS$^{\downarrow}$ \\ \hline
BungeeNeRF  & 22.21 & 0.524 & 0.428 & 19.41 & 0.401 & 0.587 \\
BungeeNeRF(adapted) & 30.42 & 0.903 & 0.194 & 24.07 & 0.770 & 0.379 \\
Mip-NeRF360  & 30.79 & 0.906 & 0.188 & 24.16 & 0.792 & 0.383 \\
Ours & \textbf{32.44} & \textbf{0.929} & \textbf{0.148} & \textbf{26.29} & \textbf{0.843} & \textbf{0.332} \\ \hline
\end{tabular}%
}

\label{tab:ablation}
\vspace{-1.5em}
\end{table*}



\end{document}